\documentclass[runningheads]{llncs}

\usepackage[utf8]{inputenc}
\usepackage[T1]{fontenc}
\usepackage[american]{babel}

\usepackage{microtype}
\usepackage[shortcuts]{extdash}
\usepackage{graphicx}
\usepackage{xcolor}
\usepackage{tabularray}
\usepackage[export]{adjustbox}
\usepackage[skins,breakable]{tcolorbox}

\usepackage[hyphens]{url}
\usepackage[colorlinks,linkcolor=blue,citecolor=blue,urlcolor=blue]{hyperref}
\usepackage{cleveref}

\begin{document}
\title{Make Literature-Based Discovery \texorpdfstring{\\}{} Great Again through Reproducible Pipelines}
\titlerunning{Bisociative Literature-Based Discovery}
\author{Bojan Cestnik\inst{1,2}\orcidID{0000-0001-8887-5706} \and Andrej Kastrin\inst{3}\orcidID{0000-0002-3495-0165} \and Boshko Koloski\inst{1,4}\orcidID{0000-0002-7330-0579} \and Nada Lavra\v{c}\inst{1}\orcidID{0000-0002-9995-7093}
}
\authorrunning{B. Cestnik et al.}
\institute{Jo\v{z}ef Stefan Institute, Ljubljana, Slovenia \\
\email{\{nada.lavrac,boshko.koloski\}@ijs.si} \\
\and
Temida d.o.o, Ljubljana, Slovenia \\
\email{bojan.cestnik@temida.si}
\and
University of Ljubljana, Ljubljana, Slovenia \\
\email{andrej.kastrin@mf.uni-lj.si} 
\and International Postgraduate School Jo\v{z}ef Stefan, Ljubljana, Slovenia
}
\maketitle
\begin{abstract}
By connecting disparate sources of scientific literature, literature\-/based discovery (LBD) methods help to uncover new knowledge and generate new research hypotheses that cannot be found from domain-specific documents alone. Our work focuses on bisociative LBD methods that combine bisociative reasoning with LBD techniques. The paper presents LBD through the lens of reproducible science to ensure the reproducibility of LBD experiments, overcome the inconsistent use of benchmark datasets and methods, trigger collaboration, and advance the LBD field toward more robust and impactful scientific discoveries. The main novelty of this study is a collection of Jupyter Notebooks that illustrate the steps of the bisociative LBD process, including data acquisition, text preprocessing, hypothesis formulation, and evaluation. The contributed notebooks implement a selection of traditional LBD approaches, as well as our own ensemble-based, outlier-based, and link prediction-based approaches. The reader can benefit from hands-on experience with LBD through open access to benchmark datasets, code reuse, and a ready-to-run Docker recipe that ensures reproducibility of the selected LBD methods. 

\end{abstract}

\section{Introduction}

In an era of rapidly increasing amounts of scientific publications, researchers face the challenge of keeping up with the latest findings, even within their narrow scientific fields. In a flood of information, researchers can overlook valuable segments of knowledge. Focused on their silos of information, researchers are incapable of delving into novel literature from neighboring domains or getting insights from possibly yet unconnected domains, as means for sparkling novel research insights or forming new hypotheses.

Literature-based discovery (LBD) is an area of natural language processing (NLP) research, with the first approaches reported in the mid-1980s~\cite{swanson1986fish}. LBD tools aim to uncover new knowledge by connecting disparate sources of literature to generate novel research hypotheses that are not evident in domain-specific documents. In his pioneering work, Swanson~\cite{swanson1986fish}, reading articles in separate literature sources on Raynaud's disease and dietary fish oil, observed that some knowledge concepts were shared between the two separate sets of documents. This led to the discovery that fish oil could be used in the treatment of Raynaud's disease, a hypothesis that was later clinically confirmed. During the last four decades, researchers have proposed several different methods, ranging from simple latent semantic indexing~\cite{gordon1998using}, more complex techniques based on knowledge graphs~\cite{sang2018sematyp}, up to novel large language models~\cite{wang2024scimon}. A detailed overview of LBD approaches can be found in studies of several authors~\cite{sebastian2017emerging,thilakaratne2019systematic,kastrin2021scientometric}.

This paper focuses on bisociative LBD, an area of research that is the focus of our research interests~\cite{lavrac2020bisociative,kastrin2021scientometric}. We explore a specific aspect of LBD that is grounded in two research traditions: bisociative knowledge discovery~\cite{berthold2012bisociative} and LBD~\cite{swanson1990medical}. By integrating these two research traditions, bisociative reasoning can support the formation of serendipitous hypotheses by discovering the connections of seemingly unrelated concepts~\cite{koestler1964art}. The main contribution of this paper is the open science approach to bisociative LBD, providing access to benchmark LBD datasets, replication of LBD discoveries, as well as a collection of Jupyter Notebooks---interactive computational narratives containing code and formatted text---that implement the main steps of the LBD process: data acquisition, text preprocessing, hypothesis discovery, and evaluation. 

The paper first explains the open science approach to data and software reuse in \Cref{sec:reusability}. To replicate the reported discoveries, \Cref{sec:data_benchmarks} provides access to several benchmark datasets used in pioneering LBD papers and more recent bisociative LBD approaches. This paper supports reproducible science through dockerized Jupyter Notebooks. Tutorials illustrating how to implement simplified LBD workflows using Jupyter Notebooks are presented in \Cref{sec:reproduced_techniques}. The developed Python notebooks demonstrate traditional LBD approaches alongside our own bisociative LBD methods, which include (i) an ensemble-based approach for identifying cross-domain bridging terms, (ii) an outlier-based method that reduces the search space of linking terms to outlier documents, (iii) a link prediction approach for ranking associations, and (iv) a large language model (LLM)-based approach to LBD domain conceptualization. The paper concludes with a discussion in \Cref{sec:discussion}.

\section{Approach to Reproducible Science}
\label{sec:reusability}

The development of modern programming languages, programming paradigms, and operating systems enabled the development of computer platforms for text analysis, and of modern integrated text analysis tools. Such platforms offer a high degree of abstraction and allow the user to focus on analyzing the results rather than on the methods used to obtain the results. 

Originally, individual algorithms were implemented as complete solutions for specific NLP problems, while second-generation NLP systems were intended to provide integrated solutions for entire NLP pipelines including text preprocessing, transformation, and text analysis tasks, ideally also with graphical user interfaces. Many of them, such as scikit-learn~\cite{pedregosa2011scikitlearn}, take advantage of operating system-independent languages, such as the Python environment, to develop complete solutions that include methods for text preprocessing and visual representation of results. These solutions also provide interfaces to call each other and thereby enable cross-solution development. 

A common practice in modern scientific publishing is to make code and data publicly available, e.g., via GitHub. Today's AI software is further confronted with the challenge of how to adapt such open science solutions to big data and deep learning, which require grid computing and GPU acceleration. Hugging Face~\cite{wolf2020transformers} exemplifies open-source collaboration within the NLP community by providing widespread access to state-of-the-art models. Similarly, ReScience~\cite{rougier2017sustainable} is a community-driven initiative that focuses on republishing and reimplementing previously published methods to promote reproducible and accessible science.

Additional means of simplifying access to NLP tools and enabling cross-solution development are modern web-based Jupyter Notebooks~\cite{toomey2017jupyter,perkel2018jupyter} that allow the creation and sharing of documents containing live code, equations, and visualizations. The open science approach to LBD implemented in this paper uses Jupyter Notebooks to practically illustrate selected concepts and LBD pipelines, allowing the reader to run the described methods on illustrative examples. The notebooks provide a set of examples that we have carefully chosen to be easy to understand but are non-trivial and can be used as recipes or starting points for developing more complex LBD scenarios. Moreover, following the recommendations and guidelines for reproducible science and FAIR benchmarking~\cite{peng2011reproducible}, we prepared dockerized versions of all methods, which provide off-the-shelf and ready-to-use environment for testing and working with the LBD methods surveyed in this paper.

\section{Benchmark Datasets Enabling Experiment Replicability}
\label{sec:data_benchmarks}

A selection of datasets prepared for reuse and tool utility verification has been constructed in a way that enables the replication of the originally reported LBD experiments. To achieve this, the query enabling to download a PubMed dataset is limited by the year of the paper presenting the originally discovered bridging terms ($b$-terms), which is different for every dataset. The summary of statistics on the datasets made available for reuse is presented in \Cref{tab:datasets}.

\begin{table}[htb]
\centering
\caption{Basic properties and statistics of domain pair datasets.}
\begin{tblr}{%
hline{1,3,9}={solid},%
hline{2}={2-4}{solid},%
width=\linewidth,%
colspec={Q[l]*{3}{Q[0.20,c]}}%
}
& \SetCell[c=3]{c} Dataset & & \\
Property & RS-DFO & Mig-Mg & Aut-CaN \\
No.\ of docs. & 1273/153 & 1156/7320 & {10,819}/5320 \\
Mean no.\ of words/doc. & 13.1/18.3 & 11.3/16.1 & 166/259 \\
Mean no.\ of terms/doc. & 2.56/1.92 & 11.5/12.1 & 4.02/4.03 \\
No.\ of unique terms & 4554/1059 & 1812/6217 & {423,153}/{316,702} \\
No.\ of common terms & 27 & 228 & {46,473} \\
No.\ of $b$-terms & 3 & 43 & 13 \\
\end{tblr}
\label{tab:datasets}
\end{table}

We used the $b$-terms, which were identified in each domain pair dataset, as the gold standard to evaluate the utility of tested LBD approaches to cross-context link discovery. \Cref{tab:b_terms} presents the $b$-terms for Raynaud's Syndrome--Dietary Fish Oil (RS-DFO), Migraine--Magnesium (Mig-Mg), and Autism--Calcineurin (Aut-CaN) domain pair datasets used in our experiments.

\begin{table}[htb]
\centering
\caption{Bridging terms for the RS-DFO, Mig-Mg, and Aut-CaN domain pairs.}
\begin{tblr}{%
hline{1,3,5,7}={solid},%
width=\linewidth,%
row{1,3,5}={belowsep=0pt},
row{2,4,6}={abovesep=0pt},
colspec={Q[1,j]}%
}
Raynaud's Syndrome--Dietary Fish Oil (RS-DFO) \\
\leftskip=1em%
\textit{Terms}: blood viscosity, platelet aggregation, vascular reactivity \\
Migraine--Magnesium (Mig-Mg) \\
\leftskip=1em%
\textit{Terms}: serotonin, spread, spread depression, seizure, calcium antagonist, vasospasm,
paroxysmal, stress, prostaglandin, reactivity, spasm, inflammatory, anti inflammatory,
5 hydroxytryptamine, calcium channel, epileptic, platelet aggregation, verapamil,
calcium channel blocker, nifedipine, indomethacin, prostaglandin e1, anticonvulsant,
arterial spasm, coronary spasm, cerebral vasospasm, convulsion, cortical spread depression, brain serotonin, \hbox{5 hydroxytryptamine} receptor, epilepsy, antimigraine,
\hbox{5 ht}, epileptiform, platelet function, prostacyclin, hypoxia, diltiazem, convulsive,
substance p, calcium blocker, prostaglandin synthesis, anti aggregation \\
Autism--Calcineurin (Aut-CaN) \\
\leftskip=1em%
\textit{Terms}: synaptic, synaptic plasticity, calmodulin, radiation, working memory, bcl 2,
type 1 diabetes, ulcerative colitis, asbestos, deletion syndrome, 22q11 2,
maternal hypothyroxinemia, bombesin \\
\end{tblr}
\label{tab:b_terms}
\end{table}

\begin{tcolorbox}[breakable,notitle,sharp corners,boxrule=0.5pt,boxsep=0pt,left=5pt,right=5pt,top=5pt,bottom=5pt]
\textbf{Novelty}: This is the first time an interested LBD researcher can get a systematic presentation and open access to the benchmark datasets and the targeted $b$-terms. The datasets are available in the \textsc{clarin.si} repository~\cite{kastrin2024clarin}.
\end{tcolorbox}

\subsection{Raynaud's Syndrome--Dietary Fish Oil (RS-DFO) Benchmark}

Swanson's~\cite{swanson1986fish} RS-DFO hypothesis linking RS and DFO literatures is the most commonly used baseline for evaluating novel LBD approaches. In our experiments, we used the following PubMed queries to build the RS-DFO dataset:
\begin{enumerate}
  \item \texttt{raynaud* [TI] AND 1900/01/01:1985/11/30 [PDAT]}
  \item \texttt{fish oil [TI] AND 1900/01/01:1985/11/30 [PDAT]}
\end{enumerate}
Note that Swanson formulated the RS-DFO hypothesis at the end of November 1985; hence, we limited the upper bound of the time window with the PubMed ``Publication Date'' (PDAT) tag to reflect the body of scientific literature available at that time. The retrieved data are stored in the dataset \texttt{swanson\_1986.psv.gz} and comprise 1273 records associated with Raynaud's disease (domain $C$) and 153 records associated with fish oil (domain $A$). The three most important terms on which we focus are ``blood viscosity'', ``platelet aggregation'', and ``vascular reactivity''.

\subsection{Migraine--Magnesium (Mig-Mg) Benchmark}

The second benchmark, the Migraine--Magnesium domain pair dataset, pertains to the discovery of implicit links between migraine ($C$) and magnesium ($A$). Swanson~\cite{swanson1988migraine} described eleven neglected connections (i.e., $b$-terms) between sets $A$ and $C$, explaining magnesium deficiency as an important cause of the onset of migraine disease in humans. In the LBD process, Swanson managed to find more than 60 pairs of articles connecting the migraine domain with magnesium deficiency through several bridging concepts. In this process, Swanson identified 43 $b$-terms connecting the two domains. For our research, we used two PubMed queries to construct the dataset:
\begin{enumerate}
    \item \texttt{migraine [TI] AND 1900/01/01:1987/12/31 [PDAT]}
    \item \texttt{magnesium [TI] AND 1900/01/01:1987/12/31 [PDAT]}
\end{enumerate}
The gathered bibliographic data is stored in the \texttt{swanson\_1988.psv.gz} dataset.

\subsection{Autism--Calcineurin (Aut-CaN) Benchmark}

The third dataset is the Autism--Calcineurin domain pair~\cite{petric2009literature}. Autism belongs to a group of pervasive developmental disorders portrayed by an early delay and abnormal development of the cognitive, communication, and social interaction skills of a person. It is a very complex, and not yet sufficiently understood domain, where precise causes are still unknown. Like Swanson~\cite{swanson1986fish,swanson1988migraine}, Petri\v{c} et al.~\cite{petric2009literature} also provided $b$-terms, 13 in total, whose importance in connecting autism to calcineurin (a protein phosphatase) was discussed and confirmed by the domain expert. The dataset was constructed using the following PubMed query:
\begin{enumerate}
    \item \texttt{autis* [TIAB] AND 1900/01/01:2007/12/31 [PDAT]}
    \item \texttt{calcineurin [TIAB] AND 1900/01/01:2007/12/31 [PDAT]}
\end{enumerate}
The retrieved PubMed data are collected in the \texttt{petric\_2009.psv.gz} dataset.

The programming scripts for preprocessing the described datasets are available in the GitHub repository of the project. The workflow involves several text-mining steps, including text tokenization, stopword removal, and word stemming/lemmatization, using the LemmaGen lemmatizer for English. Context-specific terms representing knowledge concepts are then determined using $n$-grams, which are defined as sequences of 1 to $n$ consecutive words with a minimum support frequency. The processed text is then transformed into a bag-of-words (BoW) representation using term frequency--inverse document frequency (TF-IDF) weighting. In addition, to enhance the standard workflow, terms containing words that served as query terms during document selection were removed from the dataset to ensure unbiased preprocessing.

\section{Reproduced LBD Techniques}
\label{sec:reproduced_techniques}

This section starts by introducing two novel Python pipelines, illustrating the general principle of closed LBD based on Swanson's ABC discovery approach in \Cref{sec:abc_model}, and open LBD based on Weeber's concept-based approach in \Cref{sec:open_discovery}. The section continues with brief introductions to different bisociative LBD techniques, representing prototypical approaches to closed and open LBD,\footnote{Due to space constraints, we cannot explain the differences between the open and closed discovery settings, commonly used in LBD. For details, see Weeber et al.~\cite{weeber2001using}.} explaining both the intuition behind the techniques and the implementation details. The presentation of reusable LBD pipelines, aimed at enabling replication of our own LBD approaches, includes the
(i) ensemble-based approach to identifying cross-domain bridging terms in \Cref{sec:text_based}, 
(ii) outlier-based method that reduces the search space of linking terms to outlier documents in \Cref{sec:outlier_closed}, 
(iii) outlier-based open discovery method in \Cref{sec:open_RaJoLink}, and
(iv) network-based approach to ranking associations in \Cref{sec:graph_based}.
Finally, we also present an approach to LBD domain conceptualization achieved by the AHAM LLM approach to topic modeling in \Cref{sec:aham}. \Cref{tab:reproduced_techniques} summarizes the LBD approaches implemented in Jupyter Notebooks.

\begin{table}[ht]
    \centering
    \caption{Summary of reproduced LBD techniques.}
    \begin{tblr}{%
    width=\linewidth,%
    hline{1,2,9}={solid},%
    colspec={Q[1,l]*{5}{Q[c]}}%
    }
    LBD technique & {Section \\ in paper} & {Discovery \\ mode} & Dataset & {Document \\ representation}  & {URL} \\
    Swanson's ABC~\cite{swanson1986fish} & \ref{sec:abc_model}& Closed & RS-DFO & TF-IDF BoW  & \raisebox{-0.25\totalheight}{\href{https://github.com/akastrin/ida2025lbd/blob/main/notebooks/01_closed_discovery.ipynb}{\includegraphics[height=1em,frame]{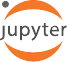}}} \\
    Concept-based~\cite{weeber2001using} & \ref{sec:open_discovery} & Open & Mig-Mg & BoW  & \raisebox{-0.25\totalheight}{\href{https://github.com/akastrin/ida2025lbd/blob/main/notebooks/02_open_discovery.ipynb}{\includegraphics[height=1em,frame]{jupyter.pdf}}} \\
    CrossBee~\cite{jursic2012cross} & \ref{sec:text_based} & Closed & Aut-CaN & TF-IDF BoW  & \raisebox{-0.25\totalheight}{\href{https://github.com/akastrin/ida2025lbd/blob/main/notebooks/03_mini_crossbee.ipynb}{\includegraphics[height=1em,frame]{jupyter.pdf}}} \\
    OntoGen-based~\cite{cestnik2017reducing}& \ref{sec:outlier_closed} & Closed & Aut-CaN  & TF-IDF BoW & \raisebox{-0.25\totalheight}{\href{https://github.com/akastrin/ida2025lbd/blob/main/notebooks/04_mini_ontogen.ipynb}{\includegraphics[height=1em,frame]{jupyter.pdf}}} \\
    RaJoLink \cite{cestnik2016exploring} & \ref{sec:open_RaJoLink} & Open & Aut-CaN  & TF-IDF BoW  & \raisebox{-0.25\totalheight}{\href{https://github.com/akastrin/ida2025lbd/blob/main/notebooks/05_mini_rajolink.ipynb}{\includegraphics[height=1em,frame]{jupyter.pdf}}} \\
    Network-based~\cite{kastrin2016link} & \ref{sec:graph_based} & Closed & RS-DFO & Network  & \raisebox{-0.25\totalheight}{\href{https://github.com/akastrin/ida2025lbd/blob/main/notebooks/06_mini_linkpred.ipynb}{\includegraphics[height=1em,frame]{jupyter.pdf}}} \\
    AHAM \& LLM~\cite{koloski2024aham} & \ref{sec:aham} &  Topics & Papers & Embeddings & \raisebox{-0.25\totalheight}{\href{https://github.com/akastrin/ida2025lbd/blob/main/notebooks/07_aham_lbd.ipynb}{\includegraphics[height=1em,frame]{jupyter.pdf}}} \\
    \end{tblr}
    \label{tab:reproduced_techniques}
\end{table}

\begin{tcolorbox}[breakable,notitle,sharp corners,boxrule=0.5pt,boxsep=0pt,left=5pt,right=5pt,top=5pt,bottom=5pt,before skip=0pt,after skip=0pt]
\textbf{Novelty}: This is the first time for an interested LBD researcher to get access to prototypical LBD techniques via tutorials implemented in dockerized Jupyter Notebooks. 
The code is available at \url{https://github.com/akastrin/ida2025lbd}.
\end{tcolorbox}

\subsection{Swanson's ABC Closed Discovery Model}
\label{sec:abc_model}

This section illustrates the general principle of LBD based on Swanson's~\cite{swanson1986fish} ABC closed discovery model, where $A$ and $C$ refer to the two domains in the $A{-}C$ domain pair, and $B$ refers to a set of potential $b$-terms. The goal is to replicate Swanson's link between Raynaud's syndrome and dietary fish oil (i.e., RS-DFO hypothesis) employing the closed discovery approach. 

For the RS-DFO hypothesis, the term ``raynaud'' represents the concept $c$, while ``fish oil'' corresponds to the concept $a$. To simulate the closed discovery mode, we exclude any records shared between domains $C$ and $A$, thereby focusing exclusively on indirect connections between the two domains. For document representation, we use article titles, while terms are represented as bigrams. Domain $C$ comprises 4554 unique terms (uni- and bi-grams), while domain $A$ contains 1059 distinct terms. We then identify intersecting terms shared between the two domains. This set consists of 27 common terms, including interesting $b$-terms such as ``blood viscosity'', ``platelet aggregation'', and ``vascular reactivity'', which were originally identified by Swanson~\cite{swanson1986fish}. Swanson's ABC closed discovery approach is illustrated in the \texttt{01\_closed\_dis\-cov\-ery.ipynb} notebook.

\subsection{Concept-Based Open Discovery}
\label{sec:open_discovery}

In this section, we extend the closed discovery approach outlined above to the open discovery mode to generate new research hypotheses. We guide the reader through reproducing Swanson's second landmark discovery, which connects migraine headaches to magnesium~\cite{swanson1988migraine}.  We aim to identify novel therapeutic candidates ($A$) for migraine headaches ($C$). 

Rather than relying on phrases extracted from plain text (as in \Cref{sec:abc_model}), this tutorial leverages a structured knowledge base, an approach originally introduced by Weeber et al.~\cite{weeber2001using}. However, in the present implementation, we use the Medical Subject Headings (MeSH) ontology~\cite{lipscomb2000medical} to facilitate the LBD process. The use of MeSH allows for structured background knowledge to reduce textual variations and synonyms of terms to standardized knowledge concepts.

Each document in the dataset is represented as a list of MeSH headings. The LBD process begins with a dataset of 1156 PubMed articles on migraines, which map to a total of 1812 unique MeSH headings. To narrow the search space, we apply a semantic type filter, restricting MeSH headings to categories such as ``Amino Acid, Peptide, or Protein'', ``Pathologic Function'' and ``Phenomenon or Process'', resulting to a more manageable set of 116 MeSH headings.

Next, we rank the filtered MeSH headings using TF-IDF and select three $b$-concepts for further investigation: ``Vasoconstriction'', ``Platelet Aggregation'', and ``Spreading Cortical Depression''. For each $b$-concept, we perform a new PubMed query to identify associated $a$-concepts. This search retrieves 2898, 6273, and 180 bibliographic records for ``Vasoconstriction'', ``Platelet Aggregation'', and ``Spreading Cortical Depression'', respectively. Applying the semantic type filter to these records yields 11, 73, and 13 unique MeSH headings for the respective $b$-concepts.

The intersection of these results contains ten candidate $a$-concepts. After applying rank aggregation~\cite{kolde2012robust}, two concepts, ``Calcium'' and ``Magnesium'', emerge as the top-ranked candidates. Notably, ``Calcium'' already has four direct connections to migraines in the literature, whereas ``Magnesium'' shows no prior joint reports, making it a promising novel hypothesis. The implementation of the concept-based open discovery approach can be found in the \texttt{02\_open\_discovery.ipynb} notebook.

\subsection{Text-Mining-Based Closed Discovery}
\label{sec:text_based}

The CrossBee method~\cite{jursic2012cross} represents documents using a BoW document representation, with TF-IDF weights used to model the importance of words and multi-word ($n$-gram) expressions. The method operates on a dataset spanning over two domains, Autism and Calcineurin, with {10,819} documents from domain Autism and 5320 documents from domain Calcineurin (\Cref{tab:datasets}). For each domain, TF-IDF weighted BoW representations are built. Next, each domain representation is aggregated, allowing for easier analysis of the text data categorized by domain. On top of the calculated statistics, several pre-prepared heuristics are used to estimate the importance of terms as bridging term candidates. These heuristics include elementary heuristics, such as \textit{freqTerm}, \textit{freqDoc}, \textit{freqRatio}, which can be combined in an ensemble heuristic for $b$-term discovery. From the the initial vocabulary, {46,473} terms that occur in both domains are selected as candidates for bridging terms and ranked according to a selected ensemble heuristic. The rank of the interesting $b$-terms from \Cref{tab:b_terms} is determined and displayed in the Receiver Operating Characteristic diagram. The Area Under the Curve is also calculated for the selected ensemble heuristic. The method is implemented in the \texttt{03\_mini\_crossbee.ipynb} notebook.

\subsection{Outlier-Based Closed Discovery}
\label{sec:outlier_closed}

Outlier-based LBD methods for closed discovery efficiently narrow down the search for cross-domain bridging terms to outlier documents, thus simplifying the exploration of bridging terms by domain experts. The workflow illustrates the experiments with the OntoGen tool~\cite{cestnik2017reducing}, which is used for $b$-term discovery by detecting outlier documents in the Autism--Calcineurin domain pair~\cite{petric2012bison,sluban2012exploring}.

The process starts with cleaning and structuring the text data, followed by feature extraction using a BoW model and a TF-IDF matrix, both of which are refined by filtering out less relevant terms. Dimensionality reduction with principal component analysis enables two-dimensional visualization of document clusters determined by $k$-means clustering. The combinations of domains and clusters are analyzed to detect overlaps and identify outlier groups (i.e., Autism [Cluster~\#1] and Calcineurin [Cluster \#0] with 356 and 45 documents, respectively). The workflow concludes with the storage of these outliers, which are saved in a file for further processing and experimentation. The illustrated method is implemented in the \texttt{04\_mini\_ontogen.ipynb} notebook.
       
\subsection{Outlier-Based Open Discovery}
\label{sec:open_RaJoLink}

RaJoLink represents an outlier-based open-discovery three-step method, applied to the Autism dataset. The method requires a predefined list of terms. In the first step (Ra), we limit the terms to MeSH headings, corresponding to semantic types ``Enzymes and Coenzymes'' and ``Amino Acids, Peptides, and Proteins''~\cite{lipscomb2000medical}, resulting in 495 terms of interest. After that, TF-IDF weighting of the terms is conducted across documents, resulting in TF-IDF BoW representation over the retrieved MeSH terms. Next, the representation is aggregated per term, resulting in a single score per term. Finally, an expert examines the terms based on their positioning over the aggregated BoW representation, selecting the most interesting ones. This manual intervention identified three terms out of 495, specifically ``calcium channel'', ``synaptophysin'', and ``lactoylglutathione'', for further investigation of their association with autism. The positions of the three selected rare terms in the sorted list of candidate rare terms were 38, 37 and 377. 

In the next step (Jo), the documents (titles and abstracts) for each rare term were retrieved from PubMed and combined into a single document list. {11,666} documents from all three domains were collected and then preprocessed similarly to step Ra. The BoW model was created and aggregated by domain. Based on the BoW model, 428 common word candidates were selected and sorted based on the TF-IDF scores. The expert selected the term ``calcineurin'', which was in position 18 of the list of 428 sorted joint word candidates.

The last step Link in RaJoLink implements the principle of closed discovery between two domains, in our case Autism and Calcineurin. Using Autism as the start domain of interest, the three-step RaJoLink LBD open discovery procedure~\cite{petric2012outlier} is illustrated in the \texttt{05\_mini\_rajolink.ipynb} notebook.

\subsection{Network-Based Closed Discovery}
\label{sec:graph_based}

The network-based approach extends the discovery of bridging terms to relation discovery, exploiting their potential to reveal bisociative relationships among concepts across different biomedical domains. The idea is to mimic the LBD discovery process as a link prediction problem~\cite{kastrin2016link}. In this tutorial, we exploit the co-citation network generated from the primary documents in the RS-DFO hypothesis. 

The initial dataset of 1426 documents from the RS-DFO experiment is expanded into a training network, including 7128 and 2072 cited references for Raynaud's disease and fish oil, respectively, retrieved from the OpenAlex database. Additionally, the testing network consists of the first 1000 cited references published after November 30, 1985. The notebook explores unsupervised link prediction using proximity measures, including common neighbors, the Jaccard coefficient, and the Adamic/Adar index. A time-sliced evaluation demonstrates high prediction accuracy, highlighting the efficacy of this approach. The approach is illustrated in the \texttt{06\_mini\_linkpred.ipynb} notebook.

\subsection{LLM-Based Conceptualization of the LBD Research Field}
\label{sec:aham}

In addition to the LBD methods presented above aimed at new hypothesis generation, we recently proposed an LLM-based approach to LBD literature conceptualization~\cite{koloski2024aham}, offering novel domain insights through topic modeling using LLMs. The method begins by using a BERTopic~\cite{grootendorst2022bertopic}, a topic modeling procedure, on initial embeddings to identify topics covering the datasets, which are subsequently named by an LLM using an expert-designed prompt. Next, the ratio of outliers to topics is calculated and multiplied by the topic name similarity between documents, forming the AHAM score. The embeddings are then fine-tuned to the corpora, and the previous step is repeated for {50,000} steps, evaluating the AHAM scores for each {10,000} steps, selecting the embedding with the lowest AHAM score. This procedure identified 19 topics of interest that represent the field of LBD. The method is implemented in the \texttt{07\_aham\_lbd.ipynb} notebook.

\section{Discussion}
\label{sec:discussion}

The concept of literate programming, introduced by Knuth in the mid-1980s~\cite{knuth1984literate}, emphasizes the importance of integrating documentation with source code, enabling both human readability and computational functionality. In recent years, reproducibility has garnered significant attention from various scientific fields and communities, with numerous studies identifying the challenges and proposing solutions to address the reproducibility crisis~\cite{baker2016scientists}. These efforts underscore the critical need for consistent practices and tools to ensure that research outputs can be independently verified and built upon.

The reproducible programming paradigm is highly relevant to LBD as it highlights the value of creating reproducible and transparent workflows that facilitate both the theoretical understanding of implemented methods and the reuse of programming code. Key issues to address include managing dependencies effectively, resolving potential environmental conflicts, such as operating system incompatibilities, and simplifying installation processes to avoid unnecessary complexity. Failure to consider these factors can hinder the ability of others to replicate results or extend the existing work. The reproducible approach has not been explicitly addressed in the literature on LBD, although recent studies have incorporated at least some demonstrative data and code segments~\cite{sebastian2017emerging,thilakaratne2019systematic}.

We assist LBD researchers by enabling the reproducibility of studies and by providing robust access to resources. These include benchmark datasets, computational notebooks, and a containerized development environment that can be effectively integrated at various stages of the research lifecycle. By offering these tools, we seek to mitigate barriers to reproducibility and foster a culture of scientific rigor and transparency within the LBD community.

In conclusion, making LBD ``great again'' involves not only advancing LBD methods but also ensuring that the underlying research practices align with the principles of reproducibility. By providing comprehensive Python notebooks wrapped in a dockerized programming environment and supported by standardized benchmark datasets, researchers can drive innovation while fostering trust and collaboration in the field.

\begin{credits}
\subsubsection{\ackname} The authors acknowledge the financial support from the Slovenian Research and Innovation Agency through the Knowledge Technologies (No.\ P2-0103), and Methodology for Data Analysis in Medical Sciences (No.\ P3-0154) core research projects, as well as Embeddings-Based Techniques for Media Monitoring Applications (No. L2-50070) research project. The young researcher grant (No.\ PR-12394) supported the work of Boshko Koloski.

\subsubsection{\discintname}
The authors have no competing interests to declare that are relevant to the content of this article.
\end{credits}

\bibliographystyle{splncs04}
\bibliography{references}

\end{document}